\documentclass[sigchi,nonacm]{acmart}

\AtBeginDocument{%
  \providecommand\BibTeX{{%
    \normalfont B\kern-0.5em{\scshape i\kern-0.25em b}\kern-0.8em\TeX}}}

\usepackage[acronym]{glossaries}

\newacronym{AI}{AI}{artificial intelligence}
\newacronym{CNN}{CNN}{convolutional neural network}
\newacronym[\glslongpluralkey={electrocardiographies}]{ECG}{ECG}{electrocardiography}
\newacronym{FOH}{FOH}{fear of hypoglycemia}
\newacronym{FBG}{heart rate monitor}{Firstbeat Bodyguard 2}
\newacronym{GVA}{consumer smartwatch}{consumer smartwatch}
\newacronym{HRV}{HRV}{heart rate variability}
\newacronym{MSE}{MSE}{mean squared error}
\newacronym{MAE}{MAE}{mean absolute error}
\newacronym{RMSE}{RMSE}{root-mean-squared error}
\newacronym{RMSSD}{RMSSD}{root mean square of the successive differences}
\newacronym{SDNN}{SDNN}{standard deviation of normal-to-normal interval}
\newacronym{T1DM}{T1DM}{type 1 diabetes mellitus}

\setcopyright{rightsretained}

\begin{document}


\title[Improving HRV Measurements from Smartwatches with Machine Learning]{Improving Heart Rate Variability Measurements from Consumer Smartwatches with Machine Learning}

\author{Martin Maritsch}
\email{mmaritsch@ethz.ch}
\affiliation{%
  \institution{ETH Zurich}
  \city{Zurich}
  \country{Switzerland}
}

\author{Caterina B\'{e}rub\'{e}}
\email{cberube@ethz.ch}
\affiliation{%
  \institution{ETH Zurich}
  \city{Zurich}
  \country{Switzerland}
}

\author{Mathias Kraus}
\email{mathiaskraus@ethz.ch}
\affiliation{%
  \institution{ETH Zurich}
  \city{Zurich}
  \country{Switzerland}
}

\author{Vera Lehmann}
\email{vera.lehmann@insel.ch}
\affiliation{%
  \institution{Inselspital, Bern University Hospital}
  \city{Bern}
  \country{Switzerland}
}

\author{Thomas Z\"uger}
\email{thomas.zueger@insel.ch}
\affiliation{%
  \institution{Inselspital, Bern University Hospital}
  \city{Bern}
  \country{Switzerland}
}

\author{Stefan Feuerriegel}
\email{sfeuerriegel@ethz.ch}
\affiliation{%
  \institution{ETH Zurich}
  \city{Zurich}
  \country{Switzerland}
}

\author{Tobias Kowatsch}
\email{tobias.kowatsch@unisg.ch}
\affiliation{%
  \institution{University of St. Gallen}
  \city{St. Gallen}
  \country{Switzerland}
}
\additionalaffiliation{
  \institution{ETH Zurich}
  \city{Zurich}
  \country{Switzerland}
}

\author{Felix Wortmann}
\email{felix.wortmann@unisg.ch}
\affiliation{%
  \institution{University of St. Gallen}
  \city{St. Gallen}
  \country{Switzerland}
}

\renewcommand{\shortauthors}{Maritsch and B\'{e}rub\'{e}, et al.}

\begin{abstract}
  The reactions of the human body to physical exercise, psychophysiological stress and heart diseases are reflected in \gls{HRV}. Thus, continuous monitoring of \gls{HRV} can contribute to determining and predicting issues in well-being and mental health.
  \gls{HRV} can be measured in everyday life by consumer wearable devices such as smartwatches which are easily accessible and affordable. However, they are arguably accurate due to the stability of the sensor. 
  We hypothesize a systematic error which is related to the wearer movement. Our evidence builds upon explanatory and predictive modeling: we find a statistically significant correlation between error in \gls{HRV} measurements and the wearer movement. 
  We show that this error can be minimized by bringing into context additional available sensor information, such as accelerometer data. This work demonstrates our research-in-progress on how neural learning can minimize the error of such smartwatch \gls{HRV} measurements.
\end{abstract}


\copyrightyear{2019} 
\acmYear{2019} 
\acmConference[UbiComp/ISWC '19 Adjunct]{Adjunct Proceedings of the 2019 ACM International Joint Conference on Pervasive and Ubiquitous Computing and the 2019 International Symposium on Wearable Computers}{September 9--13, 2019}{London, United Kingdom}
\acmBooktitle{Adjunct Proceedings of the 2019 ACM International Joint Conference on Pervasive and Ubiquitous Computing and the 2019 International Symposium on Wearable Computers (UbiComp/ISWC '19 Adjunct), September 9--13, 2019, London, United Kingdom}
\acmDOI{10.1145/3341162.3346276}
\acmISBN{978-1-4503-6869-8/19/09}




\maketitle

\glsresetall

\section{Introduction}
The ability of human heart to adapt to sudden physiological changes is visible in \gls{HRV}; the variation of subsequent inter-beat intervals across time. In fact, it informs about involuntary physiological functions, as it is a valid measure of the interaction between the autonomic nervous systems (i.e. sympathetic and parasympathetic). \gls{HRV} reflects individual reactions to physical exercise, psychophysiological stress and heart diseases. In particular, \gls{HRV} has been associated with parasympathetic activity \cite{goldberger2001relationship, von2007heart} and hypoglycemic episodes \cite{cichosz2014novel, schachinger2004increased} of type 1 diabetes mellitus patients, which are caused by a deficit in the sympathetic nervous system \cite{cryer2004diverse}.


Self-tracking devices are becoming increasingly popular: sales of sports and fitness trackers increased from 97.6 million in 2015 to 134 million in 2018 and are expected to reach 148.5 million in 2021 \cite{mordor2018smart} (cf. also \cite{henriksen2018using} for a review on availability of wrist-worn fitness wearable devices and sensors). Due to their discrete form-factor, smartwatches are suitable for continuous monitoring of the wearers vital factors. To this end, they are often equipped with measurement capabilities such as an optical heart rate sensor. Besides observing the heart rate, raw data from these sensors can as well be used to compute \gls{HRV}. Their affordability and unobtrusiveness makes smartwatches relevant to a large audience and, in particular, more likely to be adapted than professional \glspl{ECG} for measuring \gls{HRV}.


Wearables measuring \gls{HRV} can considerably contribute to improving well-being and mental health. For instance, it can predict acute complications of type 1 diabetes mellitus (i.e., hypoglycemia). As such, its benefit to better health becomes relevant to around 40 million people worldwide and this number is expected to double within the next 10 to 20 years \cite{international2017idf}.

\textbf{Objective: }
This work demonstrates research-in-progress on the potential of machine learning approaches improving smartwatch \gls{HRV} measurements. More specifically, we aim to utilize explanatory and predictive modeling to the sensor data provided by smartwatches with the goal of minimizing errors in \gls{HRV} measurements due to physical activity. 

\section{Related work}
\subsection{Consumer Wearable Devices}
Wearable devices allow non-invasive monitoring of physiological activity. In particular, smartwatches can be applied to different contexts, such as stress \cite{von2007heart}, sleep quality \cite{HRMonitoringPerformance} or physical fitness \cite{dooley2017estimating} to measure well-being through heart rate. Moreover, these devices are often equipped with a range of additional sensors, allowing for more holistic measurements. For instance, there are wrist-based consumer smartwatches that provide multiple data dimensions such as inter-beat intervals (obtained via optical sensor), three-axis accelerometer data, steps, burned calories or proprietary stress values. However, such non-professional devices need to be compared to a more precise instrument. The \emph{\gls{FBG}} is an inter-beat interval recorder employing two electrodes on the chest for measurement and that can be considered as a semi-professional device \cite{parak2013accuracy}.

Smartwatches thus collect a magnitude of data, most of which is sufficient for giving their user an overview of their daily activity. However, when being applied to more serious medical and healthcare use cases, the current measurement accuracy of wearable devices remains rather unclear \cite{el2015currently, phan2015smartwatch, ge2016evaluating, dooley2017estimating, gillinov2017variable}. 

\subsection{Heart Rate Variability}
\gls{HRV} measurements are relevant in various applications. For example, the study conducted in \cite{cichosz2014novel} used an \gls{ECG} to derive \gls{HRV} for hypoglycemia prediction and detection. However, to reach a broader audience, we propose to use consumer wearable devices such as smartwatches instead of professional \glspl{ECG} to measure \gls{HRV}.
While some studies show that data from smartwatches was highly accurate when compared with professional \glspl{ECG} in long-term measurements \cite{HRMonitoringPerformance}, other previous research observed a discrepancy between measurements from wrist-worn trackers and \glspl{ECG} \cite{dooley2017estimating, gillinov2017variable}.

\textbf{Resolution: }
\Gls{HRV} can be measured in both the time and frequency domain.
According to the Task Force of the European Society of Cardiology \cite{malik1996heart}, short-term recordings (i.e. 2 to 5 minutes) should be assessed with \gls{SDNN} and \gls{RMSSD} in the time domain.
However, \gls{RMSSD} has been found to be reliable when calculated with a sample of 10, 30 or 60 seconds, which was not always the case for \gls{SDNN}, or frequency-based measures \cite{thong2003accuracy, salahuddin2007ultra, nussinovitch2011reliability,esco2014ultra, baek2015reliability, munoz2015validity}.
Ultra-short-term measures of \gls{HRV} such as \gls{RMSSD} would thus allow monitoring streams of physiological changes with a relatively high resolution.

\textbf{\gls{HRV} and hypoglycemia: }
\gls{HRV} has been shown to correlate positively with hypoglycemia \cite{schachinger2004increased}, although in a short-term measure (i.e., 5\,min), and with continuous glucose monitoring as a valid predictor of hypoglycemia \cite{cichosz2014novel}. It is therefore in our interest to explore the possibility of using an ultra-short-term measurement of \gls{HRV} to predict hypoglycemia.




\section{Work-in-progress research}
\subsection{Problem Statement}
Many widespread smartwatches are capable of measuring \gls{HRV}, however, our analysis has shown that they incur systematic measurement errors leading to inaccurate \gls{HRV} measurements. We hypothesize that these data quality issues are related to the movement or physical activity of the wearer and the device being unable to properly measure inter-beat intervals during these times.

Smartwatches are nowadays equipped with a range of sensors relevant for health measurement, such as for example, an optical heart sensor, an accelerometer, a compass, or sensors for positioning information. These sensors can potentially provide valuable additional information.

Our aim is to minimize the error of smartwatch \gls{HRV} measurements by using additional available sensor information for correcting \gls{HRV} measurements. With our work, we want to improve the quality and reliability of the \gls{RMSSD} calculated by smartwatches. The basis of these calculations are the inter-beat intervals as measured by those devices.

\begin{figure*}
  \includegraphics[width=\textwidth]{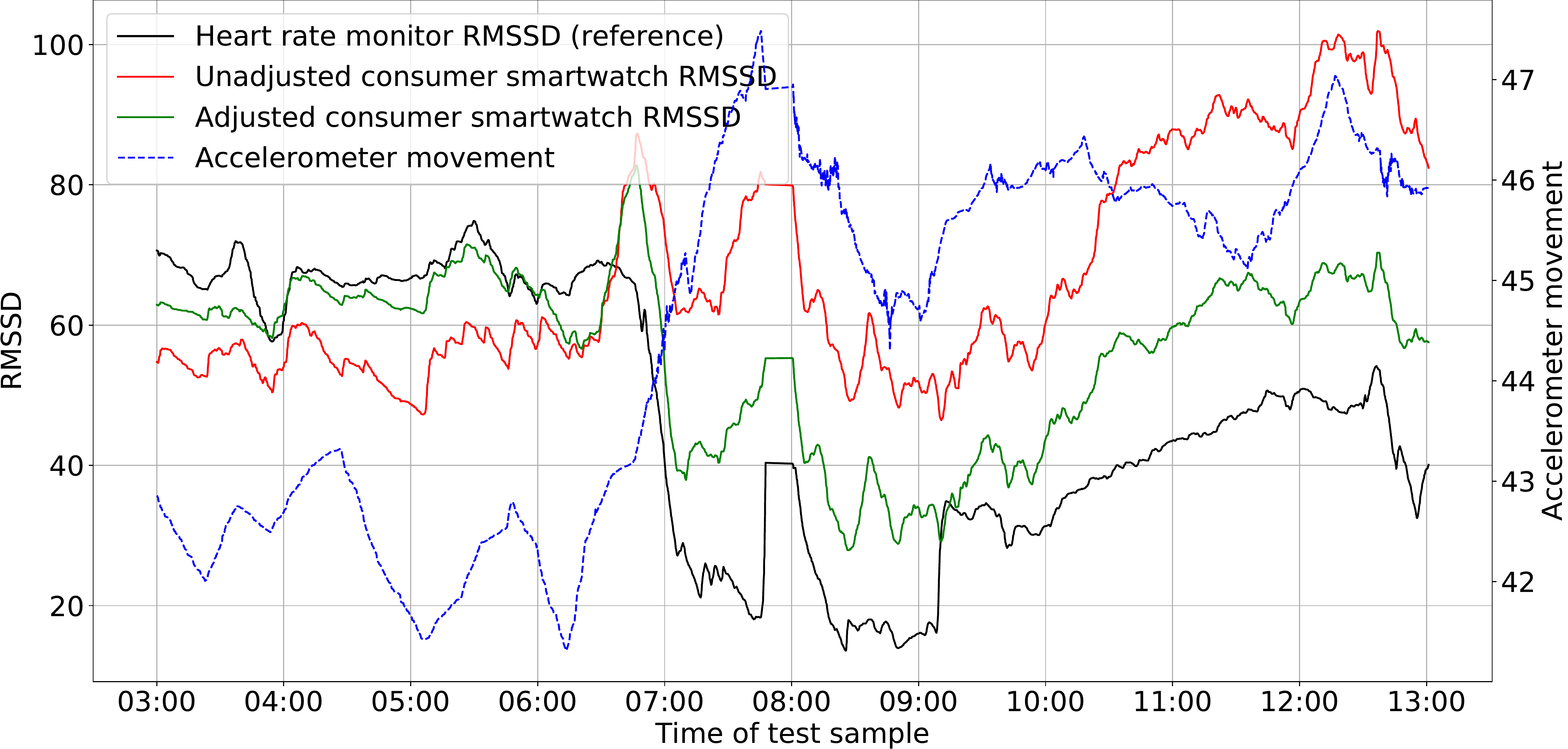}
  \caption{Samples of the \gls{RMSSD} as calculated from data of the \gls{FBG} (black, reference value) and the \gls{GVA} (red, unadjusted). The line in green shows the adjusted \gls{GVA} \gls{RMSSD}, which is the sum of the unadjusted \gls{RMSSD} and the error predicted by the neural network. The dashed blue line indicates accelerometer movement. Values have been smoothed to facilitate visualization.}~\label{fig:unadjusted-adjusted}
\end{figure*}

\subsection{Methods}
\textbf{Devices: }
In our preliminary research, we compare data obtained from a chest-based heart rate monitor (\emph{\acrlong{FBG}}) with data from one of the most widespread consumer smartwatches which is equipped with an accelerometer and an optical heart rate sensor capable of measuring inter-beat intervals.

The \gls{FBG} allows recording inter-beat intervals by means of a two-lead measurement, thus providing reference \gls{HRV} data for our investigations. Given the cited related work on smartwatches we consider obtained from the \gls{GVA} to be less accurate.
However, the \gls{GVA} provides further data dimensions, which we take into account in order to compensate for the inaccuracies in \gls{HRV} measurements imposed by inaccurate inter-beat interval data. 

For instance, we aim to improve the accuracy of \gls{RMSSD} calculations by including both inter-beat intervals and movement indices such as orientation and angular velocity data from accelerometer sensors into the corresponding calculations.

\textbf{Analysis: }
In a first step, we use explanatory analysis to compare and investigate the differences between the \gls{HRV} measurements from both devices, and how the differences relate to the wearer movement. Then, in a second step, we apply predictive modeling in order to forecast measurement errors imposed by the described wearer movements.

\textbf{Prediction Model: }
In order to estimate the aforementioned measurement error, we make use of a \gls{CNN}. \glspl{CNN} are a sub-group of neural networks, which are commonly used for classification tasks like object recognition in images \cite{krizhevsky2012imagenet}. In particular, \glspl{CNN} have also successfully been applied to medical time-series data such as for \gls{ECG} classification \cite{Zihlmann2017}.

In our model, we frame the obtained data as a time-series task in a supervised learning setting. The aim is to estimate the errors between the \gls{GVA} \glspl{RMSSD} (\emph{samples} in the machine learning model) and the reference heart rate monitor \glspl{RMSSD} (\emph{labels} in the machine learning model).

\subsection{Data}

Raw data was obtained from the \gls{FBG} via the \emph{Firstbeat SPORTS Individual} software. From the \gls{GVA}, data was gathered via a custom smartphone app utilizing the Bluetooth streaming capability of the smartwatch.

Data was recorded by a healthy individual wearing both devices simultaneously over a period of 72 hours. After pre-processing, a total of roughly 200,000 observations consisting of \gls{FBG} inter-beat interval, \gls{GVA} inter-beat interval and \gls{GVA} three-axis accelerometer data were left. We split the dataset into 80\% of the observations for training and 20\% for testing.

\section{Results}
We quantify \gls{HRV} by \gls{RMSSD} which is calculated as
\begin{equation*}
    RMSSD = \sqrt{\frac{1}{N-1}\sum_{i=1}^{N} (IBI_i - IBI_{i-1})^2}\,,
\end{equation*}
where $N$ is the number of samples within a sliding window of 60 seconds and $IBI_i$ is the $i$-th inter-beat interval within those samples.

The error of the measurements between the \gls{FBG} and the \gls{GVA} at time $t$ is defined as
\begin{equation*}
    \varepsilon_t = RMSSD_{\textrm{\gls{FBG}}_t} - RMSSD_{\textrm{\gls{GVA}}_t}\,.
\end{equation*}


\subsection{Explanatory Analysis}
We first tested for a tendency to systematic errors in HRV measurements of the \gls{GVA}, which are imposed by external influences such as the wearer movement.
As a result, we found a statistically significant positive correlation between accelerator movement and measurement error, $r = 0.22$, $p = \leq 0.001$.

\subsection{Predictive Analysis}
\textbf{Error Prediction Model: }
In our approach, we used a deep \gls{CNN} in order to predict the error of the measurement. Subsequently, the predicted error is added to the \gls{RMSSD} as calculated from the raw inter-beat intervals from the \gls{GVA} in order to get an approximation of the reference \gls{RMSSD} calculated from the \gls{FBG} data.

\textbf{Performance: }
The first experiment yielded an improvement of the \gls{RMSE} between \gls{FBG} and \gls{GVA} \glspl{RMSSD} from an initial 48.89 down to 28.50 on the test set. The statistically significant correlation of heart rate monitor and consumer smartwatch \glspl{RMSSD} in the test set was improved from $r = 0.37$, $p = \leq 0.001$ before adjusting to $r = 0.58$, $p = \leq 0.001$ after adjusting.


Figure \ref{fig:unadjusted-adjusted} visualizes this reduction of the measurement error in \glspl{RMSSD}. For the first part of the graph, the \gls{RMSSD} error is rather low. During this time, the wearer was in bed, which is reflected by low accelerometer movement. Later on, at around 06:30 when the wearer got out of bed, a rise of accelerometer movement and a deterioration of the error in raw \gls{RMSSD} measurements are observable. Additionally, it can be observed how the addition of the error as predicted by the \gls{CNN} turns from positive into negative. The sensor data suggests that a larger magnitude of accelerometer movement is related to a larger error in measurement.

\section{Conclusion} 
Prediction of \gls{HRV} measurements with smartwatches would allow non-invasive continuous monitoring of psychophysiological conditions such as the risk of hypoglycemia.
The key benefit of being able to accurately measure \gls{HRV} with smartwatches would certainly be their popularity and easiness of use in contrast to professional devices. The implemented \gls{CNN} is capable of reducing the error in \gls{GVA} \gls{HRV} measurements by taking into account additional motion information in the form of accelerometer data.


\textbf{Current Limitations: }
In future research, we will investigate the generalizability of our approach, which needs to be done in a larger-scale study. Furthermore, the restricted accuracy of the \gls{FBG} (see \cite{parak2013accuracy}) also limits the accuracy of the model evolved on top of its data. We thus aim to conduct a larger-scale study in which we gather reference data with a professional-grade \gls{ECG} device.

\textbf{Outlook: }
We found strong evidence that systematic errors in \gls{HRV} measurements from the \gls{GVA} can be minimized with the utilization of additional data by the use of neural networks.
In the future, we can generate additional data dimensions that potentially carry valuable information for our problem (e.g., the time-shifted difference of accelerometer data) by utilizing methods of feature engineering. 
While we are aware of that our proposed methods will not be capable of completely eliminating the error in smartwatch \gls{HRV} measurements, the methods we have shown contribute to more reliable measurements of physiological values such as \gls{HRV} in smartwatches. Furthermore, even more, widespread smartwatches such as the \emph{Apple Watch} should be investigated.


\begin{acks}
This work was part-funded by the Swiss National Science Foundation (SNF), Project 183569.
\end{acks}

\bibliographystyle{ACM-Reference-Format}
\bibliography{main}

\end{document}